\documentclass{llncs}
\usepackage[utf8]{inputenc}
\usepackage{amsmath,amssymb}
\usepackage{hyperref}
\usepackage{graphicx}
\usepackage{tikz}
\usetikzlibrary{arrows.meta,calc,positioning,shapes,fit}

\renewcommand{\vec}[1]{\mathbf{#1}}

\title{PoGO: A Scalable Proof of Useful Work via Quantized Gradient Descent and Merkle Proofs}
\titlerunning{PoGO v2: Proof of Gradient Optimization}
\author{José I. Orlicki}
\institute{\email{josepreprints@gmail.com}}

\usepackage{draftwatermark}

\SetWatermarkText{} 
\SetWatermarkScale{4}         
\SetWatermarkAngle{45}        
\SetWatermarkColor[gray]{0.9} 

\begin{document}
\maketitle

\begin{abstract}
We present a design called \emph{Proof of Gradient Optimization} (PoGO) for blockchain consensus, where miners produce verifiable evidence of training large-scale machine-learning models. Building on previous work~\cite{ball2017proof,bittensor,lerner2014proof}, we incorporate \emph{quantized gradients} (4-bit precision~\cite{gholami2021survey} \cite{jacob2018quantization}\cite{dettmers2022llm}) to reduce storage and computation requirements, while still preserving the ability of verifiers to check that real progress has been made on lowering the model's loss. Additionally, we employ Merkle proofs over the full 32-bit model to handle large parameter sets and to enable random leaf checks with minimal on-chain data. We illustrate these ideas using GPT-3 (175B parameters)~\cite{gpt3} as a reference example and also refer to smaller but high-performance models (e.g., \emph{Gemma~3} with 27B parameters). We provide an empirical cost analysis showing that verification is significantly cheaper than training, thanks in part to quantization and sampling. We also discuss the necessity of longer block times (potentially hours) when incorporating meaningful training steps, the trade-offs when using specialized GPU hardware, and how binary diffs may incrementally optimize updates. Finally, we note that fine-tuning can be handled in a similar manner, merely changing the dataset and the manner of sampling but preserving the overall verification flow. Our protocol allows verifiers to issue either \emph{positive} or \emph{negative} attestations; these are aggregated at finalization to either confirm the update or slash the miner.
\end{abstract}

\section{Introduction}
Traditional Proof-of-Work (PoW) blockchains (originating from Bitcoin~\cite{nakamoto2008bitcoin}) rely on miners solving cryptographic puzzles that consume large amounts of energy without producing externally useful by-products. Recent \emph{Proof of Useful Work (PoUW)} proposals~\cite{ball2017proof,bittensor} have explored turning this computational effort into tasks beneficial to society, such as protein folding or machine-learning (ML) training. However, verifying that genuine ML computations have been performed (rather than possibly forged) poses a significant challenge due to the high dimensionality of modern models and the offline nature of large-scale training.

\emph{Proof of Gradient Optimization} (\textbf{PoGO}) tackles this verification issue by requiring miners to commit to a new set of model parameters, \(\vec{\theta} \in \mathbb{R}^d\), and prove that the update reduces the model’s loss
\[
\mathcal{L}(\vec{\theta}) \;=\; \mathbb{E}_{(x,y)\in \mathcal{D}}\bigl[\ell\bigl(f_{\vec{\theta}}(x), \,y \bigr)\bigr],
\]
on a specified dataset \(\mathcal{D}\).  In practice, miners do a standard gradient-descent step
\[
   \vec{\theta}_{t+1} \;=\; \vec{\theta}_{t}\;-\;\eta \,\nabla_{\vec{\theta}}\mathcal{L}(\vec{\theta}_{t}),
\]
for learning rate \(\eta\).  The protocol relies on:
\begin{enumerate}
    \item Two-phase random verification of the \emph{quantized gradient} and sample weights from the full model.
    \item Quantized model publication (e.g., 4-bit weights) for lower-cost off-chain storage and verification~\cite{gholami2021survey,jacob2018quantization}.
    \item Merkle-based partial checks of the full 32-bit model so that large parameter sets (potentially gigabytes) need not reside entirely on-chain.
\end{enumerate}
Together, these steps enable verifiers to efficiently check training correctness at scale, even for massive models like GPT-3 or \emph{Gemma~3}. 

\paragraph{Key contributions.}  
Our approach introduces several key innovations. First, we propose \textit{quantized gradient publication}, where a 4-bit version of the model or updates is stored and transmitted off-chain. This reduces memory usage by roughly \(8\times\) while still preserving a 32-bit Merkle tree commitment for deeper verification. To ensure consistency between the quantized and full-precision models, we use \textit{Merkle-based random leaf verification}: after a publicly verifiable random delay, the network requests a single randomly chosen leaf from the full-precision model’s Merkle tree to perform an efficient integrity check.

We also include an \textit{empirical cost analysis}, estimating parameter counts and storage requirements for large models like GPT-3 (175B parameters)~\cite{gpt3} as well as smaller, high-performance models such as Gemma~3 (27B parameters). Our findings show that verification is significantly cheaper than training, supporting a secure incentive model.

To address deployment realities, we examine \textit{block time and hardware trade-offs}. Since training large models can take hours, block times may need to be longer. We analyze the cost dynamics between miners and verifiers, and show how 4-bit quantization and specialized hardware can offer up to an \(8\times\) speedup~\cite{gholami2021survey,jacob2018quantization}.

We introduce an \textit{attestation mechanism} where verifiers issue positive attestations when checks pass and negative ones when discrepancies are found~\cite{tendermint}. These votes are collected in a final aggregator block that either finalizes the update or penalizes the miner. Finally, the protocol is \textit{compatible with fine-tuning} tasks, which follow the same structure as full training with only minor adjustments.

\section{Background and Motivation}

\subsection{Large-Scale ML Models on Blockchain}
Modern large language models (LLMs) can have upwards of hundreds of billions of parameters. OpenAI’s GPT-3 has $\sim$175B parameters~\cite{gpt3}, which in 32-bit floating point can exceed 700\,GB of raw parameter data. Meanwhile, smaller but still high-performance models (e.g., \emph{Gemma~3}) might have around 27B parameters, which is significantly more compact (roughly 108\,GB in 32-bit). 

Storing or verifying such large models directly on-chain is infeasible. PoGO addresses this by using \emph{off-chain} storage (e.g., IPFS) and on-chain commitments (hashes, Merkle roots), along with random sampling to ensure correctness. Furthermore, \emph{quantized} versions of the weights (e.g., 4-bit) can reduce storage by a factor of 8, making distribution more practical~\cite{gholami2021survey,jacob2018quantization,dettmers2022llm}.

\subsection{Why Quantization?}
Quantized models replace full-precision (e.g., 32-bit float) weights with lower-precision representations such as 4-bit integers. This significantly reduces:
\begin{itemize}
    \item \emph{Memory Footprint}: $4\times$ fewer bits than 16-bit, or $8\times$ fewer than 32-bit.
    \item \emph{Bandwidth Requirements}: Cheaper to transmit off-chain or store in decentralized storage.
    \item \emph{Compute Overheads}: Specialized hardware (e.g., GPU tensor cores) can often process low-precision vectors at higher throughput, up to $8\times$ faster for 4-bit vs.\ 32-bit~\cite{gholami2021survey,dettmers2022llm}.
\end{itemize}

Hence, if PoGO requires public availability of model parameters, it makes sense to use a compact 4-bit representation. However, to preserve full precision for actual training and gradient checks, we still keep a Merkle commitment on the 32-bit model.

\section{Protocol Overview}

\begin{figure}[htbp]
\centering
\begin{tikzpicture}[
    scale=0.82, 
    transform shape,
    node distance=1.4cm,
    font=\small,
    milestone/.style={
      rectangle,
      rounded corners,
      draw,
      minimum width=5.1cm,
      minimum height=0.7cm,
      align=center,
      fill=white
    },
    arrow/.style={->, thick}
]

\node[milestone] (blockN) {Block N:\\\footnotesize - Miner trains + \\
- Publishes \texttt{hashFullModel32}, \texttt{hashQuant4}, \texttt{vrfProof}};

\node[milestone, below=1.0cm of blockN] (publish4bit) {Between N and N+(w/2):\\\footnotesize - Publish 4-bit model in IPFS\\ 
- Verifiers confirm \texttt{hashQuant4}\\
- Check quantized model’s loss};

\node[milestone, below=1.0cm of publish4bit] (blockNHalf) {Block N+(w/2):\\\footnotesize - New random seed\\
- Miner reveals random Merkle leaf\\
- Verifiers check Merkle proof};

\node[milestone, below=1.0cm of blockNHalf] (finalAtt) {After N+(w/2):\\\footnotesize - Verifiers gossip \\(positive/negative) attestations\\
- (Optional) 32-bit model publish};

\node[milestone, below=1.0cm of finalAtt] (blockNW) {Block N+w:\\\footnotesize - Aggregator tallies attestations\\
- \(\ge 2/3\) positive $\implies$ update final\\
- Otherwise, miner is slashed};

\draw[arrow] (blockN) -- (publish4bit);
\draw[arrow] (publish4bit) -- (blockNHalf);
\draw[arrow] (blockNHalf) -- (finalAtt);
\draw[arrow] (finalAtt) -- (blockNW);

\end{tikzpicture}
\caption{Timeline of PoGO v2. The miner commits to the new full model and its quantized version at block N, publishes the 4-bit model by block \(N + w/2\), and must reveal a random Merkle leaf (32-bit data) at block \(N + w/2\). Verifiers issue positive or negative attestations, which are aggregated at block \(N + w\) to finalize or slash.}
\label{fig:timeline_v2}
\end{figure}
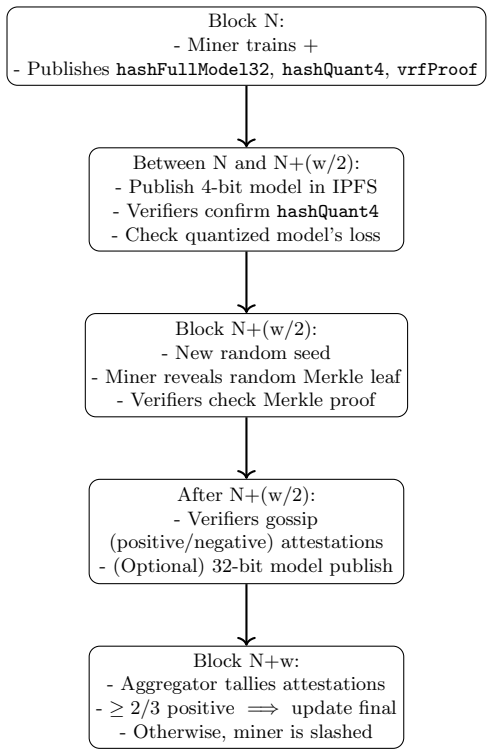

We outline PoGO’s core design in detailed form for each block at height $N$.  Let \(\theta_{t}\in \mathbb{R}^d\) represent the model parameters at iteration $t$, and suppose each block corresponds (conceptually) to an incremental update of the model using a gradient-based procedure.  The timeline is illustrated in Figure~\ref{fig:timeline_v2}, and each step is elaborated in subsequent sections.

\begin{enumerate}
    \item \textbf{Block N: Training and Commitment}
    \begin{itemize}
        \item A random miner (via VRF and based on stake) is selected to train a \emph{randomly chosen} model from the list of active tasks.
        \item The miner performs training steps (e.g., gradient descent) until it lowers the model loss \(\mathcal{L}(\theta)\) in \emph{full precision} (32-bit) more than a decrement $\epsilon$ predefined by the model owner.  Formally, it must show 
        \[
        \mathcal{L}(\vec{\theta}_{t+1}) \; < \; \mathcal{L}(\vec{\theta}_{t}) - \epsilon.
        \]
        \item The miner \emph{also} checks that the \emph{quantized} (4-bit) version of the updated weights, call it \(\widetilde{\theta}_{t+1}\), exhibits a lower loss on a small verification dataset (drawn from a VRF seed):
        \[
           \widehat{\mathcal{L}}(\widetilde{\theta}_{t+1}) 
           \;<\; 
           \widehat{\mathcal{L}}(\widetilde{\theta}_{t})
           \quad\text{on the random verification samples.}
        \]
        \item The miner constructs a Merkle tree over the full 32-bit model (potentially $\sim$GB of data). Leaves might be, for instance, 10MB each, to keep the tree size manageable (e.g., thousands or tens-of-thousands of leaves).
        \item The block includes:
        \begin{enumerate}
            \item \texttt{hashFullModel32}: The 32-bit model’s root Merkle hash.
            \item \texttt{hashQuant4}: The 4-bit quantized model’s hash.
            \item \texttt{vrfProof}: Proving the random choice of the training data mini-batch for verifying loss reduction.
        \end{enumerate}
    \end{itemize}
    \item \textbf{Between Block N and N + (w/2)}
    \begin{itemize}
        \item The miner publishes the \emph{quantized} (4-bit) model to IPFS or similar for data availability. This must happen by block $N + (w/2)$, where $w$ is the finalization window (e.g., 20 blocks).
        \item Verifiers download and verify that \texttt{hashQuant4} matches the published 4-bit model.
        \item Verifiers also confirm that on the small \emph{verification dataset} (seeded by the VRF), the quantized model indeed has lower loss than the previous model.
    \end{itemize}
    \item \textbf{Block N + (w/2): Random Leaf Challenge \& Merkle Proof}
    \begin{itemize}
        \item A new seed is derived from block \((N + w/2)\), which the miner cannot predict at block $N$.
        \item From this seed, a random leaf of the \emph{32-bit} model’s Merkle tree is selected, say index \(i\).
        \item The miner must provide the corresponding leaf data \(\texttt{leaf}_i\) and a Merkle proof linking it to \texttt{hashFullModel32}.
        \item Verifiers check the leaf contents for consistency with the previously disclosed 4-bit quantized version (up to rounding). Any mismatch implies a faked or partially inaccurate full model.
    \end{itemize}
    \item \textbf{After Block N + (w/2)}
    \begin{itemize}
        \item Each verifier issues either a \emph{positive attestation} (if checks succeed) or a \emph{negative attestation} (if a mismatch or failure is found). These are gossiped off-chain and then included in the final aggregator block at $N + w$.
        \item The miner may optionally publish the \emph{entire 32-bit model} in IPFS (or a portion, if using a sharding scheme). In some designs, only a fraction of verifiers need the full model; others might skip it if they trust the finalization.
    \end{itemize}
    \item \textbf{Block N + w: Finalization and Possible Slashing}
    \begin{itemize}
        \item A new block leader aggregates all attestations from verifiers (positive or negative).
        \item If $\ge 2/3$ of stake has signed with a \emph{positive} attestation, the block is finalized (the update is accepted).
        \item If the attestation threshold is not reached, or there are substantial negative attestations, the update is not accepted, and the miner is slashed a fraction of its stake for failing to produce a valid training step.
    \end{itemize}
\end{enumerate}

\section{Data Structures and Merkle Proofs}

\subsection{Merkle Tree for the 32-bit Model}
A typical LLM can have billions of 32-bit parameters, easily reaching gigabytes in size. We chunk these parameters into leaves of size \(\texttt{leafSize}\) (e.g., 10MB). The final Merkle tree might have thousands or tens-of-thousands of leaves. The root of this tree, \texttt{hashFullModel32}, is published on-chain in block~$N$.

\subsection{Random Leaf Reveal}
At block \((N + w/2)\), the protocol draws a \emph{public random seed} from the chain state (e.g., a VRF from block \((N + w/2)\) itself). This seed is used to choose a random leaf index $i$. The miner reveals:
\[
   \texttt{leaf}_i \quad\text{and the Merkle path from}\quad \texttt{leaf}_i \to \texttt{hashFullModel32}.
\]
Each verifier checks:
\begin{enumerate}
    \item The \(\texttt{leaf}_i\) data matches the 4-bit quantized version in the corresponding region (allowing for rounding).
    \item The Merkle path is valid.
\end{enumerate}

A single leaf check is unlikely to catch a sophisticated forgery unless the miner tries to cheat across many parameters. However, PoGO’s two-phase sampling, plus the earlier proof that the quantized model actually lowers the loss on a random mini-batch, strongly reduces any cheating probability. If a verifier finds an inconsistency, it issues a \emph{negative attestation}. To further strengthen security, the protocol could require multiple random leaves, each with a separate Merkle proof.

\section{Quantized vs.\ Full-Precision Loss Checking}

\subsection{Training in 32-bit, Verifying in 4-bit}
Miners conduct actual training (forward/backward passes) in \emph{full precision} (32-bit) to preserve numerical stability. Then they produce a 4-bit version of the updated weights. The protocol requires that this 4-bit model also shows a lower loss on a small verification dataset. This ensures that the progress is “real” even in quantized form:
\[
   \widehat{\mathcal{L}}(\widetilde{\theta}_{t+1}) \;<\; 
   \widehat{\mathcal{L}}(\widetilde{\theta}_{t})\;-\;\epsilon_\text{quant},
\]
with \(\epsilon_\text{quant}\) possibly slightly smaller than \(\epsilon\) to account for quantization error.

\subsection{Why This Works}
Since 4-bit quantization can introduce rounding noise, it is possible that small improvements in 32-bit might disappear when converted to 4-bit. However, in practice, a well-chosen quantization method preserves enough accuracy to detect genuine loss improvements, especially if the improvement is not infinitesimal~\cite{gholami2021survey,jacob2018quantization}. Additionally, a minor tolerance threshold $\epsilon_\text{quant}$ could allow for small fluctuations due to rounding.

\section{Empirical Cost Estimates: GPT-3 and Gemma~3}

\subsection{Model Sizes in Bytes}
The storage requirements for different model sizes vary depending on the number of parameters and the level of quantization. The table below summarizes the size estimates for two models—GPT-3 and Gemma~3—at 32-bit full precision and 4-bit quantization:

\begin{table}[h]
\centering
\begin{tabular}{|l|c|c|}
\hline
\textbf{Model} & \textbf{32-bit (Full Precision)} & \textbf{4-bit Quantized} \\
\hline
GPT-3 (175B parameters) & $\approx 700$ GB & $\approx 87.5$ GB \\
Gemma~3 (27B parameters) & $108$ GB & $13.5$ GB \\
\hline
\end{tabular}
\caption{Model sizes for 32-bit and 4-bit quantization schemes.}
\end{table}

Hence, the 4-bit version is significantly more compact.

\subsection{Ratio of Computation: Miners vs.\ Verifiers}

Formally, let $C_{\text{train}}$ denote the computational cost (in GPU-hours or a similar metric) for a miner to perform one gradient-descent update on a given mini-batch. This cost typically includes the forward pass, backward pass, and weight-update step in full 32-bit floating-point arithmetic:
\[
C_{\text{train}} \;=\; C_{\text{forward}}^{(32)} + C_{\text{backward}}^{(32)} + C_{\text{update}}^{(32)}.
\]

By contrast, verifiers in PoGO only need to check \emph{two} main components of the proposed block:

\begin{enumerate}
    \item \textbf{Quantized Loss Check (Forward Pass).}
    
    A verifier must confirm that the 4-bit quantized model $\widetilde{\theta}_{t+1}$ indeed achieves the claimed lower loss on a small subset of the dataset (drawn from a public VRF seed). Denote this verification subset as $\mathcal{D}_{\text{ver}} \subset \mathcal{D}$, where $\lvert \mathcal{D}_{\text{ver}} \rvert \ll \lvert \mathcal{D} \rvert$.

    The cost of this forward pass can be written as
    \[
    C_{\text{forward}}^{(4)}(\mathcal{D}_{\text{ver}}),
    \]
    where the superscript $(4)$ indicates 4-bit precision. Empirically, modern GPU tensor cores can process 4-bit representations up to $8\times$ faster than 32-bit~\cite{gholami2021survey,dettmers2022llm}. If $\alpha = \frac{\lvert \mathcal{D}_{\text{ver}} \rvert}{\lvert \mathcal{D} \rvert}$ is the fraction of the entire training data used for verification, the verifier’s forward-pass cost is approximately
    \[
    C_{\text{forward}}^{(4)}(\alpha) 
    \;\approx\; 
    \alpha \;\frac{C_{\text{forward}}^{(32)}(\mathcal{D})}{8},
    \]
    assuming the time scales linearly with the dataset size and an $8\times$ speedup going from 32-bit to 4-bit.

\item \textbf{Merkle Leaf Check.}

    The verifier must also check a \emph{random leaf} from the Merkle tree of the miner’s proposed 32-bit model. Let $C_{\text{merk}}$ denote the cost of the following operations:
    
    \begin{enumerate}
        \item retrieving a leaf value $\texttt{leaf}_i$;
        \item verifying the Merkle path;
        \item comparing the retrieved value against the quantized version.
    \end{enumerate}
    
    Since only a single (or a few) leaves are sampled, $C_{\text{merk}}$ is effectively \emph{constant} per block—independent of model size $d$ or dataset size. In practice, this cost is dominated by lightweight hash operations and a simple numerical comparison between 4-bit and 32-bit weights (allowing for rounding).
    \end{enumerate}

Hence, the total verification cost per block, $C_{\text{verify}}$, can be modeled as 
\[
  C_{\text{verify}} \;=\; C_{\text{forward}}^{(4)}(\alpha)\;+\; C_{\text{merk}}.
\]
In contrast, the miner’s cost $C_{\text{train}}$ involves the more expensive 32-bit forward and backward passes over the full mini-batch, plus other overheads (e.g., optimizer steps):
\[
  C_{\text{train}} \;\approx\; C_{\text{forward}}^{(32)}(\mathcal{D}) \;+\;
  C_{\text{backward}}^{(32)}(\mathcal{D}) \;+\;
  C_{\text{update}}^{(32)}.
\]
Because $\alpha \ll 1$ and 4-bit computations can be significantly faster than 32-bit, we typically have 
\[
   C_{\text{verify}} \;\ll\; C_{\text{train}}.
\]
For instance, even if $C_{\text{forward}}^{(32)}(\mathcal{D})$ accounts for 10 GPU-hours in training, using $\alpha = 0.01$ (just 1\% of the data for verification) and an $8\times$ speedup yields
\[
  C_{\text{forward}}^{(4)}(\alpha)
  \;\approx\;
  0.01 \times \frac{10\text{ GPU-hours}}{8}
  \;=\; 0.0125\text{ GPU-hours},
\]
plus a negligible $C_{\text{merk}}$ for the Merkle leaf check. Thus, $C_{\text{verify}}$ might easily be $\sim10$--$100\times$ cheaper than $C_{\text{train}}$, depending on the dataset fraction $\alpha$ and hardware efficiency. For very large models (billions of parameters), this gap can grow further, since \emph{most} verification overhead (the leaf-check) remains essentially constant, while the training cost scales with the model and batch size.

In short, PoGO’s design ensures that the heaviest computational burden lies on the \emph{miners} who must perform genuine training. Meanwhile, verifiers perform relatively lightweight checks on a small subset of data in 4-bit precision and confirm Merkle proofs for random leaves, thereby maintaining high security at a much lower cost.

\section{Longer Block Times}
Unlike standard blockchains with block times measured in seconds or minutes, PoGO might push block times to hours (or even a day) to accommodate meaningful training steps. This ensures:
\begin{itemize}
    \item Miners have sufficient time to do non-trivial gradient descent on large mini-batches.
    \item The network participants can download or partially download the 4-bit model from IPFS within the finalization window $w$ (e.g., 20 blocks, each possibly an hour).
\end{itemize}
Although this is a departure from fast finality blockchains, it may be acceptable in specialized ML-training blockchains where throughput is less critical than the correctness and authenticity of the training steps.

\section{Binary Diffs and Incremental Updates}
An alternative to publishing the entire model after each update is to only publish \emph{binary diffs} (the difference between consecutive 4-bit versions). This can reduce the size of each update, but for large deep learning models, the fraction of weights that change significantly in each step can still be large. Empirically, while binary diffs can yield some savings, it is often not a dramatic order-of-magnitude improvement~\cite{gholami2021survey}. Hence, it is a small optimization that can be optionally employed to reduce bandwidth and storage overhead.

\section{Fine-Tuning Use Case}
The same PoGO mechanism applies for \emph{fine-tuning} a pretrained model. The only difference is that the dataset is now user-provided or domain-specific. Verifiers still:
\begin{enumerate}
    \item Train with mini-batch provided by the model owner or paying user (possibly the entire user dataset if small).
    \item Check that the new 4-bit model lowers the loss:
    \[
    \widehat{\mathcal{L}}(\widetilde{\theta}_{\text{fine}+1}) 
    \;<\;
    \widehat{\mathcal{L}}(\widetilde{\theta}_{\text{fine}}) - \epsilon_{\text{fine}},
    \]
    on that mini-batch or a separate held-out subset.
    \item Then request a random leaf from the new full 32-bit parameters to confirm consistency via Merkle proofs.
\end{enumerate}
Since fine-tuning can overfit if the dataset is small, the random mini-batch might be the same data used for training. As long as the protocol is transparent about which data is used and the improvement is validated, it remains consistent with PoGO principles.

\section{Security Analysis}

\subsection{Two-Phase Randomness and Merkle Root}
Publishing the Merkle root at block $N$ but only revealing a random leaf at block $N + (w/2)$ prevents the miner from backdating or selectively generating a forged Merkle tree or quantized model. Because the seed for random leaf selection is only known later, the miner cannot guess which leaf to prepare with spurious data.

\subsection{Data Availability and Attestations}
Large data (4-bit or 32-bit) is stored off-chain (e.g., IPFS). The network relies on:
\begin{enumerate}
    \item \emph{Partial checks} (4-bit model hash, random leaf check).
    \item \emph{Attestations}: verifiers who can access and verify the data sign a \emph{positive} attestation if it is consistent, or a \emph{negative} attestation if they detect a mismatch (e.g., the Merkle leaf is invalid, the quantized model does not truly reduce loss, or data was not provided)~\cite{tendermint}.
\end{enumerate}
If not enough positive attestations appear by block $N+w$ (e.g., fewer than $2/3$ of stake), or if many negative attestations are submitted, the update is rejected and the miner is slashed.

\section{Incentives and Slashing}
\begin{itemize}
    \item \textbf{Rewards:} Each confirmed update yields a training reward (paid by the model owner or from protocol incentives) that goes mostly to the single random miner of that block, with a fraction (proportional to their stake) to verifiers who produce timely attestations.
    \item \textbf{Slashing:} If a miner fails to publish valid data or if sufficient negative attestations show an inconsistency, the block $N$ update is rejected at block $N+w$ and the miner’s stake is slashed.
\end{itemize}
This encourages honest participation. Because all miners who are \emph{not} the leader for block $N$ become verifiers, the system harnesses the entire staking base for robust checks. Negative attestations serve as direct evidence of misconduct, allowing the final aggregator to impose penalties.

\section{The Role of Staking and Alternative Competitive Designs}

Staking plays a critical role in this system beyond just slashing and rewards. We rely heavily on staking to enable randomness in selecting both the leader miner for each block and the next machine learning model to be trained. This randomness is essential to prevent manipulation and ensure fair participation.

One might imagine an alternative design that does not rely on staking at all. In such a system, there could be a single global model~\cite{proof-of-learning2021}, and any participant can attempt to train it. Miners would compete openly, and the winner would be the one who most successfully lowers the model’s loss function,
\[
    \mathcal{L}(\theta),
\]
on a fixed validation set. A block that claims a model \(\theta^*\) is accepted only if
\[
   \mathcal{L}(\theta^*) \;<\; \mathcal{L}(\theta_{\text{previous}}) - \epsilon,
\]
for some threshold \(\epsilon\). The ``best'' fork is then chosen as the one achieving the lowest loss so far. 

While this approach eliminates the need for randomness/staking, it also forces the protocol to focus on a single model architecture and task at a time. Our \emph{staked} PoGO design generalizes to multiple concurrent models, ensures random selection of tasks, and provides a robust mechanism for slashing dishonest updates.

\section{Comparison to Bittensor Approaches}
\label{sec:bittensor_comparison}

\paragraph{Bittensor v1: A Peer-to-Peer Intelligence Market.}
The original Bittensor framework~\cite{bittensor} proposed a decentralized “intelligence market,” wherein peers (each hosting a neural model) directly evaluate one another’s outputs and assign \emph{weights} that reflect perceived utility or “information-theoretic” value. Nodes receive additional stake if they are deemed valuable by others, creating an incentive to provide useful model outputs. This peer-to-peer approach is elegant in that it relies on local pairwise evaluations rather than a centralized oracle. However, as the Bittensor team notes, it also introduces a risk of collusion or sybil-like attacks if adversarial subgroups artificially inflate one another’s scores. Their mechanism partially mitigates this by rewarding only weights recognized by a majority share of the network, though in practice it remains challenging to definitively prove that a model’s output is \emph{genuinely} useful beyond local evaluations.

\paragraph{Bittensor v2: Stake-Based Consensus for Utility Scoring.}
A subsequent version of Bittensor~\cite{bittensor} delves deeper into an incentive function that combines network “consensus” checks with pairwise utility signals. This version formalizes a two-team (honest vs.\ cabal) game to analyze how stake evolves under different weight-assignment strategies. The design aims to penalize nodes that assign obviously skewed weights by limiting their overall stake growth if a majority of honest nodes disagree. While this improves collusion resistance compared to simple local weighting, the protocol’s complexity increases, and it still relies on collectively agreeing that certain weights are “correct” or “incorrect” in an inherently subjective, model-to-model sense.

\paragraph{Critique and Comparison.}
Both Bittensor versions share the vision of transforming costly computations (in their case, model inference or representational learning) into a decentralized marketplace of \emph{machine intelligence}. However, \textbf{PoGO} differs in scope and methodology by providing a \emph{cryptographically verifiable} proof of actual \emph{training progress}, rather than primarily relying on peer-based agreement about outputs’ quality. In Bittensor, a node’s “usefulness” is established by neighbors’ feedback signals, which can be manipulated if dishonest peers coordinate. Meanwhile, PoGO enforces a more direct measure of work: each block must present evidence that it has lowered a publicly verifiable loss on a known dataset.  

We view these lines of research as complementary rather than directly competing. Bittensor’s approach excels at building an \emph{ecosystem} of diverse models that can discover and reward novel capabilities, whereas PoGO is well-suited for \emph{verifying large-scale training} (e.g.\ fine-tuning or full model updates) using quantized gradients and Merkle proofs. Future hybrid systems might incorporate Bittensor-style local scoring within PoGO’s cryptographic framework to more richly reward high-performing sub-networks while still preserving strong security against collusion.

\section{Additional Incentive Mechanisms:\\Storage Rental and Dynamic Price Governance}
\label{sec:incentives_storage_price}

While Sections 9--10 describe basic rewards and slashing, PoGO also incorporates a \emph{storage rental} mechanism to ensure that models do not linger cost-free on the network.  In particular, uploading a new model or fine-tuning dataset requires users to commit a certain amount of \emph{POGO} tokens to rent storage for some number of blocks.  If this rental expires, nodes need not store or attest to that model.

\subsection{Renting Storage for Models}
When a user issues an \texttt{UploadModel} transaction, they:
\begin{itemize}
    \item Provide a \emph{hash} of the model (and possibly partial data).
    \item Commit \emph{POGO} tokens to cover \(\texttt{rentedBlocks}\) of storage.
    \item Within a \emph{modelUploadWindow}, they must upload the model to IPFS or a similar network.  If they fail to do so, verifiers can attest that the model is unavailable, and the system rejects it.
\end{itemize}
A user can \emph{top up} storage later (e.g.\ via \texttt{TopupStorageRental}) if training or forking continues beyond the initially funded period.  If no top-up is provided by the time storage expires, other nodes may discard the model data.

\subsection{Consensus-Driven Price Adjustments}
PoGO maintains a \emph{dynamic} per-block price for storage (per GB per block) and for compute (per training step).  We call these:
\[
\texttt{gigaPrice},\quad
\texttt{basicComputePrice}.
\]
Each consensus leader can \emph{nudge} these prices by a small $\pm \Delta$ (e.g.\ up to 0.01\% per block), capturing supply-demand fluctuations.  Thus, if network usage surges, leaders will collectively trend the price upward; if usage declines, it trends downward.  This dynamic governance ensures a rough market equilibrium without external oracles.

\subsection{Forking and Fine-Tuning Fees}
Any user can fork an existing model (\texttt{ForkModel} transaction) by paying a new round of storage rental.  Fine-tuning tasks also require committing POGO tokens to pay for \emph{compute usage}, with a separate fraction of \texttt{basicComputePrice} possibly discounted (e.g., a \emph{fineTuningFraction} parameter if the dataset is smaller).

\subsection{Incentive Impact}
This approach aligns each participant’s economic incentives:
\begin{itemize}
    \item \textbf{Model Uploaders} pay for both storage time and future training steps; if they fail to provide data, they lose their deposits.
    \item \textbf{Miners (Block Leaders)} earn block rewards \emph{plus} fees from performing the gradient optimization, scaled by \texttt{basicComputePrice}.
    \item \textbf{Consensus (Stakeholders)} collaboratively adjusts storage/compute prices in small increments, ensuring the network remains neither overburdened nor underpriced.
\end{itemize}
Overall, these dynamics encourage stable, long-term use of PoGO for large-scale training or fine-tuning tasks, with \emph{deferred publication} and \emph{rental top-ups} ensuring data availability only so long as it is economically justified.

\section{Conclusion and Future Directions}

We have presented \textbf{PoGO}, enhancing earlier \emph{Proof of Gradient Optimization} with several key ideas. These include quantized models (4-bit)~\cite{gholami2021survey,jacob2018quantization} to drastically reduce bandwidth and compute overhead for verification, and a Merkle proof of the full-precision model to ensure no hidden tampering with high-precision parameters. We introduced positive and negative attestations, enabling quick rejection (and slashing) if verifiers detect dishonest updates. Our empirical cost estimates demonstrate that verifying large models—such as GPT-3 or Gemma~3—is viable, with verification costs an order of magnitude lower than training. We also acknowledged extended block times as a design trade-off to accommodate meaningful ML training, and ensured fine-tuning compatibility with minimal changes to the verification pipeline.

Future work includes exploring advanced \emph{zero-knowledge proofs} for partial gradient verification and refining quantization strategies for even greater efficiency. \textbf{PoGO} opens up new horizons where blockchain security and real-world ML training can reinforce each other, providing a decentralized ecosystem for building, verifying, and sharing large models.

\bigskip

\section*{Acknowledgments}
We thank the broader blockchain and AI communities for early discussions, and the reviewers of earlier PoGO drafts for foundational concepts.

\appendix

\section{Basic Formal Model and Proof Sketches}
\label{appendix:formal-pogo}

This appendix presents a high-level formalization of the PoGO protocol and sketches of the main security properties. We focus on four properties: 
\begin{enumerate}
    \item \textbf{Authentic Training:} Miners must actually perform valid gradient descent steps that reduce model loss (rather than fabricating updates). 
    \item \textbf{Merkle Integrity:} Miners cannot tamper with 32-bit parameters after committing the Merkle root. 
    \item \textbf{Data Availability:} The quantized model and necessary parameter leaves must be published and verifiable (or the block is slashed). 
    \item \textbf{Slash for Dishonesty:} If a mismatch or failure is detected (e.g., an inconsistent Merkle leaf), the dishonest miner is slashed.
\end{enumerate}

\subsection{System Model and Actors}

\paragraph{Participants.}
\begin{itemize}
    \item A set of \emph{miners/stakers} $\{\mathcal{M}\}$, each with:
        \begin{enumerate}
            \item A \emph{stake} of tokens locked on-chain, subject to slashing for dishonest behavior.
            \item Sufficient computational resources (GPU, etc.) to perform gradient descent on large-scale models.
        \end{enumerate}
    \item A set of \emph{verifiers} (which may be the same stakers or an overlapping subset of them) who download the \emph{quantized model} and partially verify that the miner’s block is correct. Each verifier issues either a \emph{positive} or \emph{negative} attestation.
    \item A \emph{model owner} (potentially distinct from miners) who uploads or sponsors a model/dataset for training. This actor pays the training reward, sets acceptable parameters, etc. In some cases, this can be the entire community.
\end{itemize}

\paragraph{Data Structures.}
\begin{itemize}
    \item \(\theta_t \in \mathbb{R}^d\): The \emph{full-precision} (32-bit) model parameters at iteration $t$.
    \item \(\widetilde{\theta}_t \in \mathbb{Q}^{d}\): A \emph{quantized} (4-bit) representation of \(\theta_t\). Each parameter is reduced to 4 bits (with some rounding).
    \item A \emph{Merkle tree} $\mathcal{T}(\theta_t)$ with leaves that chunk the full 32-bit parameter array. The Merkle root $H(\theta_t)$ is included in the block as \texttt{hashFullModel32}.
\end{itemize}

\paragraph{Blockchain and Timing.}
We assume a blockchain with discrete block heights $N \in \{0,1,2,\dots\}$. The protocol often uses an extended \emph{finalization window}, $w$, to give verifiers time to check data availability, perform a random leaf challenge, and produce attestations.

\subsection{Protocol Definition (Simplified)}

At block $N$, a randomly chosen \emph{miner} (via VRF or stake-weighted lottery) proposes an update from $\theta_{t}$ to $\theta_{t+1}$ with the following steps:

\begin{enumerate}
    \item \textbf{Compute Full-Precision Update:} 
    \[
       \theta_{t+1} \;=\; \theta_{t} \;-\; \eta \,\nabla_{\theta} \mathcal{L}(\theta_{t}),
    \]
    for some mini-batch of the training dataset. Check that $\mathcal{L}(\theta_{t+1}) < \mathcal{L}(\theta_{t}) - \epsilon$ (for a protocol-defined $\epsilon > 0$).

    \item \textbf{Quantize:}
    \[
       \widetilde{\theta}_{t+1} \;=\; \mathrm{Quant4}(\theta_{t+1}),
    \]
    to produce a 4-bit representation. The miner tests that $\widetilde{\theta}_{t+1}$ still achieves a lower (approx.) loss on a small \emph{verification subset} $\mathcal{D}_{\mathrm{ver}}$ drawn from a VRF seed.

    \item \textbf{Merkle Commit:}
    \begin{itemize}
        \item Build a Merkle tree $\mathcal{T}(\theta_{t+1})$ from the 32-bit array.
        \item Let $H(\theta_{t+1})$ be its root. Publish this root as \texttt{hashFullModel32}.
        \item Also publish $\texttt{hashQuant4} \leftarrow H(\widetilde{\theta}_{t+1})$ (or a simple hash of the 4-bit array).
        \item Include a \texttt{vrfProof} for the random mini-batch selection.
    \end{itemize}

    \item \textbf{Data Availability:}
    \begin{itemize}
        \item By block $N + (w/2)$, the miner must \emph{upload} $\widetilde{\theta}_{t+1}$ (the 4-bit model) to IPFS or a similar system so that verifiers can download it and check $\texttt{hashQuant4}$.
        \item Each verifier runs a forward pass on $\widetilde{\theta}_{t+1}$ and ensures it yields a lower loss than $\widetilde{\theta}_t$ (within $\epsilon_{\mathrm{quant}}$ tolerance).
    \end{itemize}

    \item \textbf{Random Leaf Challenge (Block $N + (w/2)$):}
    \begin{itemize}
        \item A fresh \emph{random seed} $r$ is derived from block $(N + w/2)$. 
        \item A random leaf index $i \leftarrow \{1, \dots, L\}$ is chosen (where $L$ is the number of leaves in the Merkle tree).
        \item The miner reveals $\texttt{leaf}_i$ and the Merkle path proving $\texttt{leaf}_i$ is consistent with $H(\theta_{t+1})$. 
        \item Verifiers check that $\texttt{leaf}_i$ matches the corresponding portion of $\widetilde{\theta}_{t+1}$ (allowing for 4-bit rounding).
    \end{itemize}

    \item \textbf{Attestation and Finalization (Block $N + w$):}
    \begin{itemize}
        \item Verifiers issue \emph{positive} or \emph{negative} attestations.
        \item If fewer than a threshold (e.g., $2/3$ of stake) provide positive attestations, or if negative attestations are sufficient to prove a mismatch, the miner is \emph{slashed} and the block is rejected.
        \item Otherwise, $\theta_{t+1}$ is finalized, and the miner collects a reward.
    \end{itemize}
\end{enumerate}

\subsection{Adversarial Model}

\begin{itemize}
    \item An adversary may control a subset of miners, attempting to:
    \begin{enumerate}
        \item Publish a \emph{forged} update that does not actually reduce the model loss.
        \item Provide inconsistent data in the 4-bit model or the random leaf (to save computation or cheat).
        \item Omit data entirely (data unavailability).
    \end{enumerate}
    \item However, the adversary is bound by the random selection of leaves, the VRF-chosen verification subset, and the \emph{eventual finalization} which aggregates honest verifiers’ attestations.
\end{itemize}

\subsection{Key Properties and Sketch Proofs}

\subsubsection{Property~1: Authentic Training (No Fabrication of Loss Reduction)}
\begin{description}
    \item[\textbf{Definition.}] A block $B_N$ \emph{fraudulently claims} a loss reduction if the proposed $\theta_{t+1}$ does \emph{not} actually yield $\mathcal{L}(\theta_{t+1}) < \mathcal{L}(\theta_{t}) - \epsilon$, yet the miner tries to pass off the block as valid.
    \item[\textbf{Proof Sketch.}] 
    \begin{enumerate}
        \item The VRF-chosen mini-batch $\mathcal{D}_{\mathrm{ver}}$ for verifying $\widetilde{\theta}_{t+1}$ is only revealed to the miner shortly before block publication. Thus, the miner cannot pre-compute a “spoofed” model that looks good on an unknown batch unless it actually trains or guesses extremely well.
        \item Each verifier runs a forward pass on $\widetilde{\theta}_{t+1}$ to check the claimed improvement. If it is not significantly better than $\widetilde{\theta}_{t}$, a negative attestation results.
        \item Any attempt to “fake” improvement across many verifiers is highly improbable unless real training occurs, given that $\mathcal{D}_{\mathrm{ver}}$ is random and verifiers each compute the forward pass themselves. 
        \item Therefore, with high probability, the adversary cannot repeatedly fabricate loss reductions without eventually being caught and slashed.
    \end{enumerate}
    \end{description}

\subsubsection{Property~2: Merkle Integrity (No Post-Hoc Tampering)}
\begin{description}
    \item[\textbf{Definition.}] Once a miner publishes $\texttt{hashFullModel32} = H(\theta_{t+1})$ in block $N$, the \emph{Merkle integrity} property states that the miner cannot \emph{later} present a different 32-bit model $\theta'_{t+1}$ that has a different leaf (or leaves) but the same Merkle root.
    \item[\textbf{Proof Sketch.}]
    \begin{enumerate}
        \item By the collision-resistance of the Merkle tree’s underlying hash function, forging a different leaf that yields the same root is infeasible.
        \item At block $N + (w/2)$, the random leaf challenge reveals a leaf index $i$. The miner must present $\texttt{leaf}_i$ and the Merkle path to $H(\theta_{t+1})$.
        \item If the miner had stored an inconsistent leaf in $\theta_{t+1}$, or tries to swap in a different leaf, it would cause a mismatch in the Merkle path or in the 4-bit rounding. Verifiers detect this immediately and issue negative attestations.
    \end{enumerate}
    Hence, no post-hoc tampering is possible once the root is published.
    \end{description}

\subsubsection{Property~3: Data Availability (4-bit Model + Leaf Sampling)}
\begin{description}
    \item[\textbf{Definition.}] \emph{Data availability} holds if a sufficiently large subset of verifiers can (1) download the 4-bit model $\widetilde{\theta}_{t+1}$, and (2) check the randomly requested 32-bit leaf. Without data availability, verifiers cannot evaluate the claimed update.
    \item[\textbf{Proof Sketch.}]
    \begin{enumerate}
        \item The protocol enforces that if the miner does \emph{not} publish $\widetilde{\theta}_{t+1}$ by block $N + (w/2)$, honest verifiers cannot perform checks, thus must cast a negative attestation for “unavailable data.” 
        \item If negative attestations pass a certain threshold, the block is slashed and the update is invalid.
        \item Similarly, if the miner refuses to reveal the random leaf or reveals it incorrectly, verifiers submit negative attestations. 
    \end{enumerate}
    Consequently, the miner has a strong economic incentive to provide all required data in a timely manner.
    \end{description}

\subsubsection{Property~4: Slash for Dishonesty (Enforcement)}
\begin{description}
    \item[\textbf{Definition.}] \emph{Slash for dishonesty} means that if the miner publishes contradictory data (e.g., inconsistent leaves, inflated loss reductions) or refuses to provide data, they lose a stake deposit.
    \item[\textbf{Proof Sketch.}]
    \begin{enumerate}
        \item The final aggregator block $N + w$ collects attestations. If a critical mass of verifiers (e.g., $\ge 2/3$ stake) refuses to attest positively, the chain’s consensus rules mark the update as \emph{failed} and slash the block proposer.
        \item Given that a majority of verifiers are honest (standard assumption), any \emph{provable} contradiction in the update or data unavailability leads to negative attestations from a large honest set, triggering the slash condition.
        \item This penalty ensures rational miners only propose updates they can back with valid training computations and consistent data, preserving protocol correctness.
    \end{enumerate}
    \end{description}

\subsection{Additional Security Discussion}
\paragraph{Probabilistic Nature of Checks.}
Because (i) the mini-batch for verifying loss improvements and (ii) the index for the random leaf challenge are both chosen by a \emph{VRF seed} after the miner has partially committed, any attempt at forging or selective omission is caught with probability close to $1$ over repeated attempts.

\paragraph{Costs of Collusion.}
If an adversarial miner tries to bribe or collude with enough verifiers to produce fraudulent \emph{positive} attestations, it needs to control $>1/3$ or $>1/2$ stake (depending on the final threshold).  Such an adversary could, in principle, pass off fake updates.  But that is essentially a \emph{Byzantine majority} scenario, where no blockchain protocol can guarantee security if a majority is dishonest.  Hence, the protocol’s standard assumption is that a majority (or supermajority) of stake remains honest.

Our PoGO formal model codifies how \emph{quantized gradient descent} updates, Merkle commitments, random leaf challenges, and distributed attestations together enforce honest training. The proofs capture the essential argument: \emph{to pass repeated checks, a miner must genuinely improve the model’s loss and remain consistent in both 4-bit and 32-bit parameter sets.} Any inconsistency or refusal to provide data triggers negative attestations and slashing, thereby protecting the protocol from fabricated updates or ephemeral tampering.


\begin{thebibliography}{00}

\bibitem{ball2017proof}
Ball, M., Rosen, A., Sabin, M., Vasudevan, P.:
\newblock Proofs of Useful Work.
\newblock \emph{Electronic Colloquium on Computational Complexity}, 2017.

\bibitem{bittensor}
Bittensor Project:
\newblock \url{https://bittensor.com} (accessed 2025-03-17).

\bibitem{lerner2014proof}
Lerner, S.:
\newblock Proof of unique blockchain storage revised.
\newblock \url{https://bitslog.com/2014/11/03/proof-of-local-blockchain-storage/}, 2014.

\bibitem{tendermint}
Kwon, J.:
\newblock Tendermint: Consensus without mining.
\newblock \url{https://tendermint.com/static/docs/tendermint.pdf}, 2014.

\bibitem{gpt3}
Brown, T. et al.:
\newblock Language Models are Few-Shot Learners.
\newblock \emph{Advances in Neural Information Processing Systems (NeurIPS)}, 2020.

\bibitem{nakamoto2008bitcoin}
Nakamoto, S.:
\newblock Bitcoin: A Peer-to-Peer Electronic Cash System.
\newblock \url{https://bitcoin.org/bitcoin.pdf}, 2008.

\bibitem{gholami2021survey}
Gholami, A., Kim, S., Dong, Z., et al.:
\newblock A Survey on Quantization Methods for Efficient Neural Network Inference.
\newblock \emph{arXiv preprint} arXiv:2103.13630, 2021.

\bibitem{jacob2018quantization}
Jacob, B., Kligys, S., Chen, B., et al.:
\newblock Quantization and Training of Neural Networks for Efficient Integer-Arithmetic-Only Inference.
\newblock \emph{Proc. CVPR}, 2018.

\bibitem{dettmers2022llm}
Dettmers, T., Zettlemoyer, L., et al.:
\newblock 8-bit Optimizers via Block-wise Quantization for Large Language Models.
\newblock \emph{arXiv preprint} arXiv:2208.07339, 2022.

\bibitem{proof-of-learning2021}
Jia, J., Wang, T., Gong, N.Z., et al.:
\newblock Proof-of-Learning: Definitions and Practice.
\newblock \emph{NeurIPS}, 2021.

\end{thebibliography}
\end{document}